\documentclass{article}
\usepackage{makecell} 
\usepackage{amssymb}
\usepackage{lineno}
\usepackage{graphicx}
\usepackage{wrapfig}
\usepackage{hyperref}
\usepackage{url}
\usepackage{amsmath, amsthm, amsfonts}
\usepackage{enumitem}
\usepackage{dsfont}
\usepackage[ruled, noend]{algorithm2e}
\usepackage{mathrsfs}
\usepackage{multicol}
\usepackage{multirow}
\usepackage{minitoc}
\usepackage{booktabs}
\usepackage{subcaption}
\usepackage{caption}
\usepackage[dvipsnames]{xcolor}
\usepackage[most]{tcolorbox}

\usepackage{listings}

\usepackage[preprint]{corl_2025} 

\newcommand{\qa}[1]{{\textcolor{cyan}{\textbf{#1}}}}

\title{CrashAgent: Crash Scenario Generation \\ via Multi-modal Reasoning}

\author{
  Miao Li$^1$, Wenhao Ding$^2$, Haohong Lin$^1$, Yiqi Lyu$^3$,\\ 
  \textbf{Yihang Yao$^1$, Yuyou Zhang$^1$, Ding Zhao$^1$} \\
  $^1$Carnegie Mellon University\ \ \ $^2$NVIDIA Research\ \ \ $^3$Northwestern University \\
  \texttt{limiao@andrew.cmu.edu} \\
}

\begin{document}
\maketitle


\begin{abstract}    
Training and evaluating autonomous driving algorithms requires a diverse range of scenarios. However, most available datasets predominantly consist of normal driving behaviors demonstrated by human drivers, resulting in a limited number of safety-critical cases. This imbalance, often referred to as a long-tail distribution, restricts the ability of driving algorithms to learn from crucial scenarios involving risk or failure -- scenarios that are essential for humans to develop driving skills efficiently.
To generate such scenarios, we utilize Multi-modal Large Language Models to convert crash reports of accidents into a structured scenario format, which can be directly executed within simulations.
Specifically, we introduce CrashAgent, a multi-agent framework designed to interpret multi-modal real-world traffic crash reports for the generation of both road layouts and the behaviors of the ego vehicle and surrounding traffic participants.
We comprehensively evaluate the generated crash scenarios from multiple perspectives, including the accuracy of layout reconstruction, collision rate, and diversity. The resulting high-quality and large-scale crash dataset will be publicly available to support the development of safe driving algorithms in handling safety-critical situations.

\end{abstract}


\section{Introduction}
\label{sec:intro}

\begin{wrapfigure}{r}{0.5\textwidth} 
\vspace{-5mm}
\centering
\includegraphics[width=1.0\linewidth]{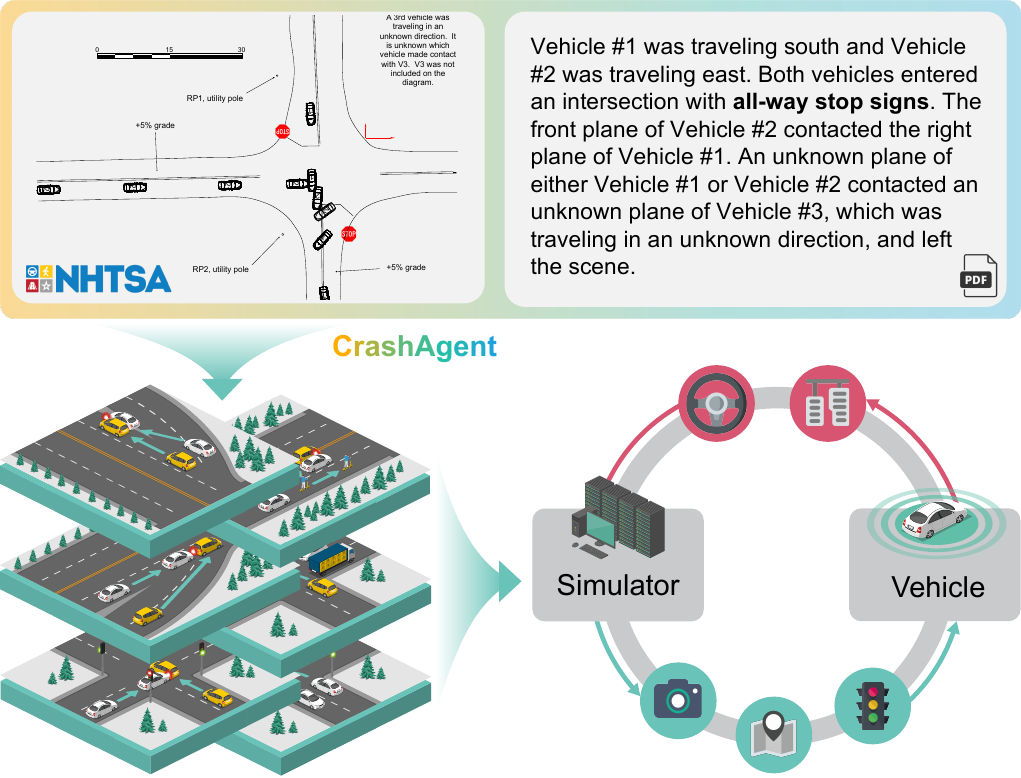}
\vspace{-5mm}
\caption{CrashAgent the crash report and diagram to scenarios that can be tested in simulators.}
\label{fig:teaser}
\vspace{-5mm}
\end{wrapfigure}

The evaluation of autonomous vehicles (AVs) heavily relies on the diversity of test scenarios. However, real-world driving scenarios exhibit a long-tail distribution~\cite{ding2025surprise}, where rare and complex events -- particularly those critical to safety -- occur infrequently but are crucial for performance assessment. Consequently, the generation of safety-critical scenarios~\cite{ding2023survey} plays a vital role in identifying potential failure modes and enhancing AV safety before deployment.

Current approaches to generating safety-critical scenarios, frequently grounded in adversarial attack techniques~\cite{hanselmann2022king, wang2021advsim}, encounter two main limitations. First, the generated scenarios often lack realism, deviating from plausible or naturalistic driving behavior. Second, these methods tend to explore only a limited portion of the behavioral space, resulting in low scenario diversity. These constraints hinder their effectiveness in capturing the wide and unpredictable range of events that AVs may face in real-world environments.

Recent progress in visual language models (VLMs)~\cite{zhang2024vision} presents a promising avenue to address these limitations. Trained on extensive multi-modal datasets, VLMs exhibit strong reasoning abilities that integrate visual and linguistic information~\cite{zhang2024improve, chen2024visual}. These models are capable of interpreting and describing complex interactions between road users and their environments. By utilizing their capacity to understand and reconstruct safety-critical situations, VLMs offer a foundation for generating scenarios that are both diverse and semantically meaningful.

However, even state-of-the-art VLMs are still not inherently capable of imagining corner driving cases without external guidance. To facilitate the exploration of infrequent yet critical scenarios, we incorporate real-world crash reports from the National Highway Traffic Safety Administration (NHTSA) as a reference. These reports offer detailed, human-annotated descriptions and structured diagrams of crash incidents, supplying the contextual grounding necessary for VLMs to generate more realistic and diverse scenarios.

In this paper, we introduce CrashAgent, a novel multi-agent framework for generating crash scenarios that combines visual reasoning with tool integration (shown in Figure~\ref{fig:teaser}). CrashAgent is composed of three key components: (1) the Sketch Agent, which produces high-level visual summaries of sketch diagrams using visual tree-of-thought prompting~\cite{chen2024visual}; (2) the Road Agent, which translates these summaries into plausible road layouts and traffic contexts; and (3) the Scenario Agent, which transforms the textual summaries into detailed, simulation-ready scripts.

\textbf{Statement of Contributions.} This work makes the following key contributions:
\vspace{-2mm}
\begin{itemize}[leftmargin=0.2in]
    \item We introduce a novel crash scenario generation method that leverages visual tree-of-thought prompting to enhance the reasoning and planning capabilities of VLMs.
    \item We perform a thorough evaluation demonstrating that the scenarios generated by our framework significantly outperform baselines in terms of both diversity and realism. 
    \item A large-scale dataset of safety-critical scenarios derived from the NHTSA crash reports is to be released, supporting future research and benchmarking in autonomous driving safety.
\end{itemize}

\section{Related Works}
\label{sec:related_work}

\textbf{Safety-critical and Crash Scenarios}\quad 
Dashboard camera is a common source for recording traffic crash scenarios, and such data has been systematically compiled into video datasets, including MM-AU~\cite{fang2024abductive} and Car Crash Dataset~\cite{bao2020uncertainty}. However, these datasets are limited to front-view visual data and lack annotations such as road maps and vehicle trajectories, which are essential for training driving algorithms.
A promising approach to addressing this limitation involves generating safety-critical scenarios through adversarial perturbations of existing datasets~\cite{ding2023survey, ransiek2024adversarial}, particularly those with comprehensive annotations like nuScenes~\cite{caesar2020nuscenes}, nuPlan~\cite{caesar2021nuplan}, and the Waymo Open Motion Dataset~\cite{ettinger2021large}. One intuitive strategy is to apply trajectory optimization to maximize collision rate as an objective. For instance, KING~\cite{hanselmann2022king} incorporates differentiable vehicle dynamics as constraints to enhance scenario realism. Further improvements are seen in~\cite{yin2024regents}, which address issues such as implausible diverging paths and unavoidable collisions. FREA~\cite{chen2024frea} enhances scenario feasibility by integrating the largest feasible region into the optimization process.
Another research direction employs Reinforcement Learning (RL) to manipulate vehicle behaviors, attacking the ego vehicle~\cite{ding2021multimodal, ransiek2024goose, zhang2023learning, wang2024efficient}. For example, \cite{wang2024efficient} introduces a Prior Risk Estimation Model as a reward function. AdvSim~\cite{wang2021advsim} applies black-box optimization techniques to transform existing scenes into safety-critical scenarios. To enhance scenario diversity, \cite{huang2024cadre} employs an evolutionary algorithm based on the Quality-Diversity framework~\cite{pugh2016quality} to explore the generation space. Despite their efficiency, adversarial approaches often result in scenarios with constrained realism and limited behavioral diversity.

\textbf{LLM for Scenario Generation}\quad
LLMs~\cite{chang2024survey} have demonstrated strong capabilities in following instructions. Consequently, recent studies have increasingly integrated LLMs into scenario generation. For instance, LCTGen~\cite{tan2023language} employs GPT to transform scenario description language into vector representations, which are then used to train a conditional generative model for simulating agent behaviors. Building on this approach, \cite{wang2025enhancing} demonstrates that scenarios generated by LCTGen can enhance the safety performance of neural planners.
DriveGen~\cite{zhang2025drivegen} utilizes LLMs in a two-stage pipeline: the first stage generates maps and vehicle assets, and the second employs a VLM to produce trajectories guided by selected waypoints from the initial output. Additionally, LLMs are utilized as constraint generators in simulations. CTG++~\cite{zhong2023language}, for example, translates human preferences into cost functions to guide inference within diffusion models. Expanding on this, \cite{liu2024controllable} incorporates a chain-of-thought mechanism to enhance the cost function generation process.
Recent efforts~\cite{mei2025llm, ruan2024traffic, lu2024multimodal} have also investigated controlling generation outcomes through natural language inputs, often in combination with retrieval-augmented generation (RAG). Although most scenarios are generated in vector space, high-fidelity rendering has also been explored. For example, \cite{zhang2024chatscene} uses LLMs to convert primitives from retrieved examples into SCENIC~\cite{fremont2019scenic} scripts for execution in Carla. Similarly, \cite{rubavicius2024generating} enables scenario updates via conversational inputs, which are then rendered in Carla.
Despite the integration of RAG, these approaches primarily rely on direct user inputs and basic instructions and thus tend to focus on generating typical or relatively simple scenarios.

\textbf{Scenario Generation from Crash Report Conversion}\quad 
Crash reports from real-world accidents can serve as valuable references for generating safety-critical scenarios. Numerous datasets are available online, such as the NHTSA Crash Investigation Sampling System~\cite{nhtsa2025ciss}, the NYC Motor Vehicle Collisions~\cite{nyc2025}, and the California Autonomous Vehicle Collision Reports~\cite{avcollision2025}. In~\cite{gao2025risk}, structured scripts of safety-critical scenarios are analyzed, while \cite{miao2024dashcam} employs a VLM to transform dashboard camera video into SCENIC scripts for reproducing accident scenarios in CARLA. SoVAR~\cite{guo2024sovar} converts crash reports into scenarios based on pre-defined road maps, which are then used to evaluate autonomous vehicle systems. Using a similar data pipeline, \cite{zhang2025accidentsim} further transforms the reports into 3D scenarios. Additionally, \cite{lu2024multimodal} and \cite{cai2025text2scenario} utilize text descriptions from NHTSA crash reports to reconstruct both road maps and behavioral aspects within the scenarios. Although these studies demonstrate the feasibility of converting crash reports into safety-critical scenarios, they have only tested a limited number of examples and primarily rely on LLMs to extract basic information from natural language. In this paper, we extend this approach by leveraging the multimodal reasoning capabilities of VLMs and present our findings on a comprehensive dataset.

\begin{figure}[t]
    \centering
    \includegraphics[width=1.0\linewidth]{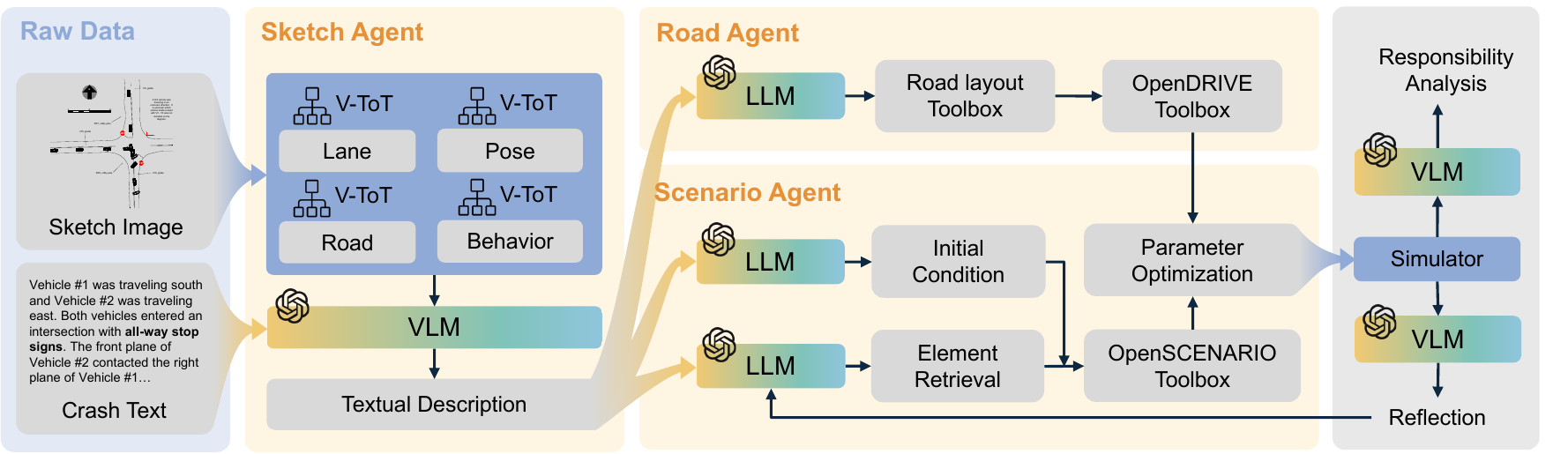}
    \caption{Detailed framework of CrashAgent}
    \label{fig:framework}
\vspace{-6mm}
\end{figure}

\vspace{-2mm}
\section{Methods} 
\label{sec:methods}
\vspace{-2mm}

To convert the multi-modal crash reports, Crash Agent first extracts spatial information and behavior from the multimodal report (\textbf{Sketch Agent}) and then reconstructs the road network (\textbf{Road Agent}) and scenarios (\textbf{Scenario Agent}). Details of the Crash Agent framework are shown in Figure~\ref{fig:framework}. We will introduce these agents in the following sections. 

\vspace{-2mm}
\subsection{Sketch Agent}
\vspace{-2mm}

Sketch Agent interprets the sketch image in the report to provide a textual description, which will be used by Road Agent to generate road networks and by Scenario Agent to generate dynamic scenarios. For road networks, we leverage VLM to recognize the number of roads, the connection relationship between roads, the number of lanes, and the corresponding traffic direction. For vehicles and other road users, VLM is used to extract not only the position of key waypoints shown in the sketch images, but also maneuver behaviors or events, for example, lane change and collision. 

Accurate information about roads and road users depends on a spatial understanding of elements within sketch images, a capability that even leading VLMs such as GPT-4o and Claude-3.7 have been reported to lack~\cite{yan2023inherent}.
Therefore, we propose a visual tree-of-thought (V-ToT) prompting method and a voting strategy to solve this problem. 
The prompting approach first breaks down the task into distinct subtasks that represent the different aspects of the scenario's properties. The VLM is required either to process a basic sketch image containing fewer components or to handle a complex sketch image supplemented by a detailed auxiliary textual description. 
The Sketch Agent then utilizes the output of V-ToT and returns a final description of the road layout and scenario events with a major voting strategy. 


\subsubsection{Visual Tree-of-Thought Prompting}

Since VLMs have difficulty in recognizing spatial attributes such as the number of lanes and the locations of road users, we propose visual tree-of-thought (V-ToT) prompting, which is motivated by two key observations: 
(1) VLMs exhibit improved spatial reasoning when processing images with fewer visual components, and (2) directing the attention to a specific region of interest (ROI) helps VLMs produce more accurate outputs.

Our method organizes sketch images into four hierarchical visual prompting trees, corresponding to a different aspect of the scenario: (1) Road Identification Tree estimates road layout based on the estimated road boundaries; (2) Lane Identification Tree captures the number and location of lanes in the road network; (3) Road User Location Tree detects the locations of road users relative to roads and lanes; (4) Road User Behavior Tree identifies the movements between two key waypoints. 
Each tree has three levels: the low-level nodes contain images with a small ROI and number of components, while the high-level nodes aggregate the regions and components from the images of all the child nodes. 
The design of the four visual trees is further explained as follows. See more implementation details and examples in the supplementary materials. 

\textbf{Road Identification Tree}\quad The leaf nodes contain pairs of estimated road boundaries with all other components removed. The VLM is tasked with identifying both the location and direction of each road boundary. The root node presents a single image containing all estimated road boundaries, where the VLM is required to summarize the total number of roads and their relative positions. 

\textbf{Lane Identification Tree}\quad This tree is used to determine the number of lanes within each road segment. Leaf nodes display the regions between two detected lane segments, while middle-level nodes represent the areas between pairs of estimated road boundaries. The root node visualizes the entire road network, providing global context for lane estimation.

\textbf{Road User Location Tree}\quad This tree is designed to identify key waypoints of road users. Each leaf node displays a single segmentable region. In the middle layer, the tree enumerates all key waypoints, with each node focusing on one specific waypoint and its corresponding road segment. At the root node, the VLM is queried to analyze the entire image, including both the road network and the positions of all road users.

\textbf{Road User Behavior Tree}\quad Given the complexity of behavior analysis, we introduce two trees. The first tree extends the structure of the Road User Location Tree by presenting two sequential key waypoints in each low-level node to capture temporal dynamics such as lane changes or turns. The second tree consists of two levels: each leaf node displays all key waypoints of a single road user, while the root node aggregates the key waypoints and trajectories of all road users.

\subsubsection{Voting with Priority}

The aforementioned V-ToT prompting decomposes the complex spatial reasoning task into a series of manageable subtasks. Once all subtasks are completed, we feed both their outputs and the original sketch image into a VLM to generate a comprehensive scene description, encompassing detailed road layouts, road user positions, and behaviors. 
However, since these trees independently produce multiple outputs related to layout, location, and behavior, inconsistencies may arise across their root-level responses. To resolve such conflicts and align the scene understanding among different trees, we apply a priority-based aggregation strategy. 
Specifically, road layouts inferred from the Road Identification Tree are prioritized over those from the Lane Identification Tree. 
Similarly, lane layouts from the Lane Identification Tree are prioritized over the lane-related details inferred by the Road User Location Tree. 
Guided by these hierarchical priorities, the Sketch Agent can resolve the conflict and generate a final interpretation of the sketch image, which serves as input for both the Road Agent and the Scenario Agent.

\vspace{-2mm}
\subsection{Road Agent}
\vspace{-2mm}

The Road Agent converts textual descriptions of road layouts into the standardized OpenDRIVE format. Instead of relying on the VLM to directly generate fully parameterized OpenDRIVE scripts, we utilize its capability to operate with predefined toolkits, emulating the way humans typically conceptualize and describe road networks.

We categorize road layouts into three primary types: single roads, intersections, and interchanges incorporating ramps. For each category, we define structural parameters such as lane count, number of connecting branches, curvature, and ramp merging or diverging configurations. The Road Agent initially employs these tools to construct a high-level representation of the road network, and subsequently calls lower-level functions to produce the corresponding OpenDRIVE scripts.

\vspace{-2mm}
\subsection{Scenario Agent}
\vspace{-2mm}

The Scenario Agent is tasked with converting behavioral descriptions of agents into the standardized OpenSCENARIO format. Although VLMs are capable of generating syntactically correct scripts, they often fail to produce reliable parameter values that accurately capture the critical risk factors detailed in crash reports.
To mitigate this shortcoming, we introduce an intermediate representation layer between the raw textual input and the final script output. 
This intermediate layer consists of pre-defined scenario elements and numerical parameter placeholders tied to key factors of the scenario.
The Scenario Agent first constructs a scenario template using these elements, and then applies a parameter optimization module to assign concrete values to the placeholders. This process yields an OpenSCENARIO script that not only accurately reconstructs the crash event but also reflects the essential risk factors identified in the original report.

\vspace{-2mm}
\subsubsection{Scenario Element Retrieval}
\vspace{-2mm}

Based on the statistical analysis of crash scenarios in the dataset, we pre-defined 42 scenario elements to serve as foundational components. A detailed list of these elements is provided in Appendix.
The Scenario Agent selects from these elements to construct a scenario template, drawing on inputs provided by the Sketch Agent and Road Agent.
Specifically, the required information includes the following:
\textbf{Road Network}: the OpenDRIVE format generated from the Road Agent;
\textbf{Initial Position}: the road and lane locations of agents;
\textbf{Intention}: the intended maneuvers of the agents during the incident, such as going straight, turning (left/right), reversing, or executing a U-turn;
\textbf{Behavior Changes}: modifications in agent behavior during the event, including both single-agent actions (e.g., lane or speed changes) and multi-agent interactions (e.g., recognizing potential collisions and initiating evasive maneuvers).



\vspace{-2mm}
\subsubsection{Numerical Parameter Optimization}
\vspace{-2mm}

After constructing the scenario template, the Scenario Agent optimizes the learnable parameters within the selected elements to align with the provided crash description. A genetic algorithm is employed to find appropriate numerical values with the following optimization objectives:
(1) ensuring that the minimum distance between road users involved in a collision is sufficiently small;
(2) maintaining a sufficiently large minimum distance between road users not involved in a collision; and
(3) preserving the temporal sequence of multiple crash events as described in the input.
During optimization, relevant constraints must be enforced. For instance, in a rear-end collision scenario, the initial positions of the vehicles must satisfy spatial constraints with realistic following distances.

\vspace{-2mm}
\subsubsection{Enrich Scenario with Additional Agents}
\vspace{-2mm}

To enhance the complexity of the generated scenarios, the Scenario Agent can also introduce non-player-character (NPC) road users. For each NPC, we prompt the LLM to generate initial positions, incorporate additional scenario elements specific to the NPC, and search for corresponding numerical parameters. The optimization objective for NPCs is to ensure that their trajectories maintain a safe distance from all other entities in the scenario.

\begin{figure}[t]
\centering
\includegraphics[width=0.93\linewidth]{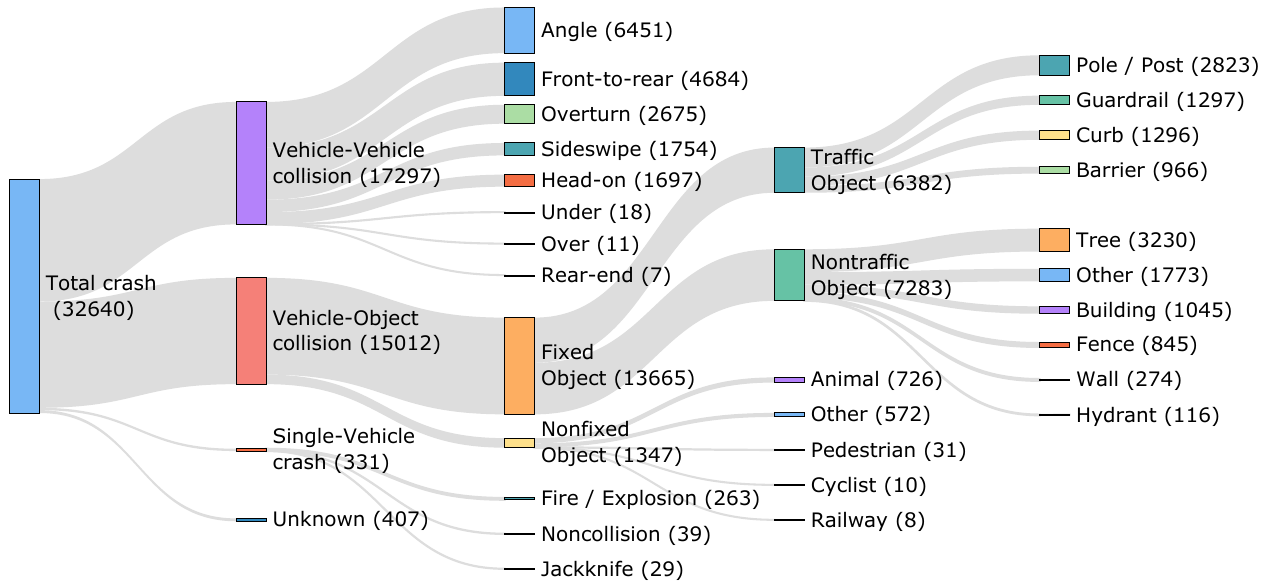}
\vspace{-1mm}
\caption{A hierarchical category of all crash scenarios in the NHTSA CISS crash dataset.}
\vspace{-4mm}
\label{fig:category}
\end{figure}

\vspace{-2mm}
\section{Experiments}
\label{sec:exp}
\vspace{-2mm}

Our experiments try to answer the following research questions. \qa{Q1.}~Do all the modules in CrashAgent improve generation performance? \qa{Q2.}~Does CrashAgent generate feasible scenarios that exhibit the risky factor from the real-world data? \qa{Q3.}~Does CrashAgent identify the initiator of the critical scenario and the responsibility according to traffic regulations?
To answers these questions, we show quantitative ablation studies and qualitative evaluation on scenario reconstruction and novel scenario generation with NPC vehicles.
Additionally, we further show examples of using CrashAgent to identify the road users at fault and valid the analysis via counterfactual scenario generation. 

\vspace{-2mm}
\subsection{Experiment Setting}
\vspace{-2mm}

\textbf{Dataset Preparation}\quad 
We utilize the crash reports from the NHTSA CISS website~\cite{nhtsa2025ciss}, which comprise 32,640 samples with accident diagrams and corresponding textual descriptions of the crash events. Through a thorough analysis of the samples, we categorize the crash events into a hierarchical structure, as shown in Figure~\ref{fig:category}, which encompasses vehicle-vehicle collisions, vehicle-object collisions, and single-vehicle crashes. More details about the dataset cleaning and pre-processing can be found in the supplementary.

To evaluate the performance of CrashAgent on real-world data, we randomly select 150 samples from the NHTSA dataset.
We further create 100 synthetic road layouts, namely 22 single roads with randomly sampled lane organization and curvature (Single Road), and 78 intersections with three to five road branches and one to five lanes on each side (Intersection (Easy)). We also augment the 78 intersection cases with dense patterns to test the robustness of the tree-building step in V-ToT (Intersection (hard)).

\textbf{Baselines}\quad
We evaluate three state-of-the-art VLMs, including GPT-4o, Gemini-1.5-flash, and Claude 3.7. We compare their performance with the 3 different modules of CrashAgent. 

\textbf{Metrics} \quad
We use the following metrics: Acc, ratio of accurate pairs of lane number; $\text{Err}_{\text{lane}}$, mean error of the number of lanes on the wider side of the road; $\text{Err}_{\text{lane-1}}$, mean error with 1-tolerance over the number of lanes on the wider side; $\text{Err}_{\text{\#road}}$, mean error of number of roads; $\text{Err}_{\text{abs}}$, sum of error over number of lanes on both side; $\text{Err}_{\text{sum}}$, mean error of total number of lanes; $\text{Acc}_{\text{\#road user}}$, ratio of generated scenarios with accurate number of road users; collision rate, ratio of generated scenarios containing crash. More details and more experimental results can be found in the Appendix.

\vspace{-2mm}
\subsection{Ablation Study of Three Modules in CrashAgent (\qa{R1})}
\vspace{-2mm}

\textbf{Sketch Agent}\quad We leverage the synthetic dataset to validate whether Sketch Agent helps with the identification of road layout and lane arrangement. 
The results in Table~\ref{tab:sketch_agent_ablation} show that with the Sketch Agent, all three VLMs achieve higher accuracy and lower error rates. 
For simple scenarios, Gemini achieves the best performance, but for complex scenarios, GPT-4o and Claude are better.
However, we observe that Claude does not always benefit from Sketch Agent, especially in hard intersection scenarios. 
The difference between the easy and hard groups lies in the quality of the visual tree. With a dense pattern compromising the image segmentation process, the Road Identification Tree and Lane Identification Tree contain unnecessary nodes. Despite the quality of visual trees, Sketch Agent still improves the accuracy of road and lane identification, indicating that even without a precise dismemberment of the image, V-ToT still improves the spatial reasoning ability of VLM. 

\begin{table}[!ht]
    \vspace{-4mm}
    \caption{Evaluation results on synthetic dataset with different VLMs.}
    \label{tab:sketch_agent_ablation}
    \vspace{1mm}
    \centering
    \scalebox{0.8}{
    \begin{tabular}{c|c|c|c|c|c|c|c|c}
        \toprule
        Scenario & VLM & Sketch Agent & Acc ($\uparrow$) & $\text{Err}_{\text{lane}}$ ($\downarrow$) & $\text{Err}_{\text{lane-1}}$ ($\downarrow$) & $\text{Err}_{\text{\#road}}$ ($\downarrow$) & $\text{Err}_{\text{abs}}$ ($\downarrow$) & $\text{Err}_{\text{sum}}$ ($\downarrow$) \\
        \midrule
        \multirow{6}{*}{Single Road} 
        & \multirow{2}{*}{GPT-4o} 
        & w/o  & 0.040 & 1.880 & 0.960 & \textbf{0.000} & 2.760 & 2.080  \\
        & & w/ & \textbf{0.360} & 1.080 & 0.480 & \textbf{0.000} & 2.200 & \textbf{0.040}  \\
        \cline{2-9}
        & \multirow{2}{*}{Gemini}  
        & w/o & \rule{0pt}{2.3ex}0.100 & 1.300 & 0.480 & 0.060 & 2.360 & 1.480   \\
        & & w/ &                 0.180 & \textbf{0.620} & \textbf{0.100} & 0.220 & \textbf{1.720} & 0.960   \\
        \cline{2-9}
        & \multirow{2}{*}{Claude}  
        & w/o  & \rule{0pt}{2.3ex}0.082 & 0.857 & 0.143 & \textbf{0.000} & 2.163 & 1.224  \\
        & & w/ &                  0.260 & 1.520 & 0.820 & \textbf{0.000} & 3.080 & \textbf{0.040}  \\

        \midrule 
        \multirow{6}{*}{\makecell[c]{Intersection\\(Easy)}} 
        & \multirow{2}{*}{GPT-4o} 
        & w/o & 0.293 & 1.043 & 0.364 & 0.468 & 2.093 & 2.093 \\ 
        & & w & 0.407 & 0.753 & 0.210 & 0.696 & 1.457 & 1.037 \\ 
        \cline{2-9}
        &  \multirow{2}{*}{Gemini} 
        & w/o & \rule{0pt}{2.3ex}0.382 & 0.820 & 0.216 & 0.456 & 1.562 & 1.562   \\
        & & w/ &                 \textbf{0.582} & \textbf{0.515} & \textbf{0.121} & 0.500 & \textbf{1.048} & 1.012   \\
        \cline{2-9}
        & \multirow{2}{*}{Claude} 
        & w/o & \rule{0pt}{2.3ex} 0.479 & 0.682 & 0.161 & \textbf{0.443} & 1.364 & 1.364   \\
        & & w/ &                 0.373 & 1.237 & 0.610 & 0.706 & 2.271 & \textbf{0.678}   \\
        
        \midrule
        \multirow{6}{*}{\makecell[c]{Intersection\\(Hard)}} 
        & \multirow{2}{*}{GPT-4o} 
        & w/o  & 0.270 & 0.989 & 0.259 & 0.620 & 1.900 & 1.900   \\   
        & & w/ & \textbf{0.500} & 0.643 & 0.226 & \textbf{0.478} & 1.179 & 1.036   \\
        \cline{2-9}
        & \multirow{2}{*}{Gemini} 
        & w/o  & \rule{0pt}{2.3ex}0.273 & 0.974 & 0.247 & 0.658 & 1.948 & 1.948   \\
        & & w/ &                  0.250 & 1.119 & 0.369 & 0.578 & 2.237 & 2.050   \\
        \cline{2-9}
        & \multirow{2}{*}{Claude} 
        & w/o  & \rule{0pt}{2.3ex}0.475 & \textbf{0.540} & \textbf{0.014} & 0.595 & \textbf{1.169} & 1.169   \\
        & & w/ &                  0.400 & 0.967 & 0.417 & 0.647 & 1.983 & \textbf{1.017}   \\
        \bottomrule
    \end{tabular}}
    \vspace{-3mm}
\end{table}

\textbf{Road Agent}\quad
To evaluate the influence of using Road Agent on the recognition of the road network, we compare the OpenDRIVE script generation result between Road Agent and a plain VLM on 100 samples from the NHTSA dataset. The results shown in Table~\ref{tab:road_agent_ablation} indicate that our Road Agent dramatically improves the reconstruction accuracy of the road network. 

\begin{table}[!ht]
    \vspace{-4mm}
    \caption{Evaluation results on NHTSA dataset with and without Road Agent.}
    \label{tab:road_agent_ablation}
    \vspace{1mm}
    \centering
    \scalebox{0.8}{
    \centering
    \begin{tabular}{c|c|c|c|c|c|c}
    \toprule
        Road Agent & Acc ($\uparrow$) & $\text{Err}_{\text{lane}}$ ($\downarrow$) & $\text{Err}_{\text{lane-1}}$ ($\downarrow$) & $\text{Err}_{\text{\#road}}$ ($\downarrow$) & $\text{Err}_{\text{abs}}$ ($\downarrow$) & $\text{Err}_{\text{sum}}$ ($\downarrow$) \\
        \midrule
        w/o & 0.407 & 0.812 & 0.532 & 0.74  & 1.392 & 1.134 \\
        w/ &\textbf{0.495} & \textbf{0.500} & \textbf{0.317} & \textbf{0.27} & \textbf{1.102} & \textbf{0.995}\\
        \bottomrule
    \end{tabular}}
    \vspace{-2mm}
\end{table}

\textbf{Scenario Agent}\quad
We compare the OpenSCENARIO script generation capability between our Scenario Agent and a plain VLM on 50 samples from the NHTSA dataset and summarize the results in Table~\ref{tab:scenario_agent_ablation}. We find that Scenario Agent improves the metrics for evaluating the crash of the scenario. 

\begin{table}[!ht]
    \vspace{-4mm}
    \caption{Evaluation results on NHTSA dataset with and without Scenario Agent.}
    \label{tab:scenario_agent_ablation}
    \vspace{1mm}
    \centering
    \scalebox{0.81}{
    \centering
    \begin{tabular}{c|c|c|c|c}
    \toprule
        Scenario Agent & Script Completeness
                        & Collision Rate ($\uparrow$) 
                        & $\text{Acc}_{\text{\#road user}}$ ($\uparrow$) 
                        & $\text{Acc}_{\text{initial relative position}}$ ($\uparrow$) 
                         \\
        \midrule
        w/o & \text{not guaranteed} & 0.08 &  0.90 & $<0.2$   \\
        w/  & \textbf{\text{guaranteed}} & \textbf{0.92} & \textbf{0.98} &  $>0.6$  \\ 
        \bottomrule
    \end{tabular}}
    \vspace{-3mm}
\end{table}

\vspace{-2mm}
\subsection{Qualitative Evaluation on NHTSA Crash Report (\qa{R2}, \qa{R3})}
\vspace{-2mm}

We now provide a qualitative evaluation of CrashAgent on the NHTSA crash dataset. We first evaluate the reconstruction quality of crash scenarios and then validate whether the model can generate novel scenarios based on the given crash report. Finally, we ask CrashAgent to identify the responsibility of the accident and then generate a counterfactual scenario that avoids the collision.

\begin{figure}[t]
\centering
\includegraphics[width=1.0\linewidth]{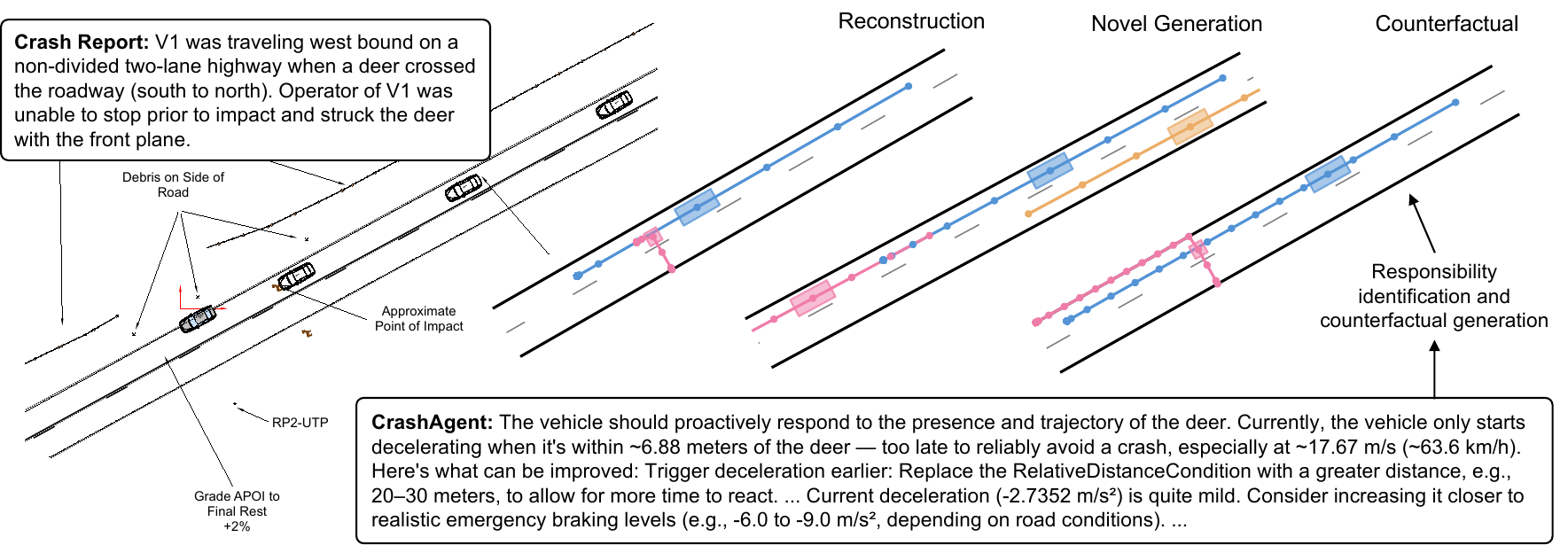}
\vspace{-3mm}
\caption{Generation results from CrashAgent with a crash between a vehicle and a deer.}
\vspace{-2mm}
\label{fig:example1}
\end{figure}

\begin{figure}[t]
\centering
\includegraphics[width=1.0\linewidth]{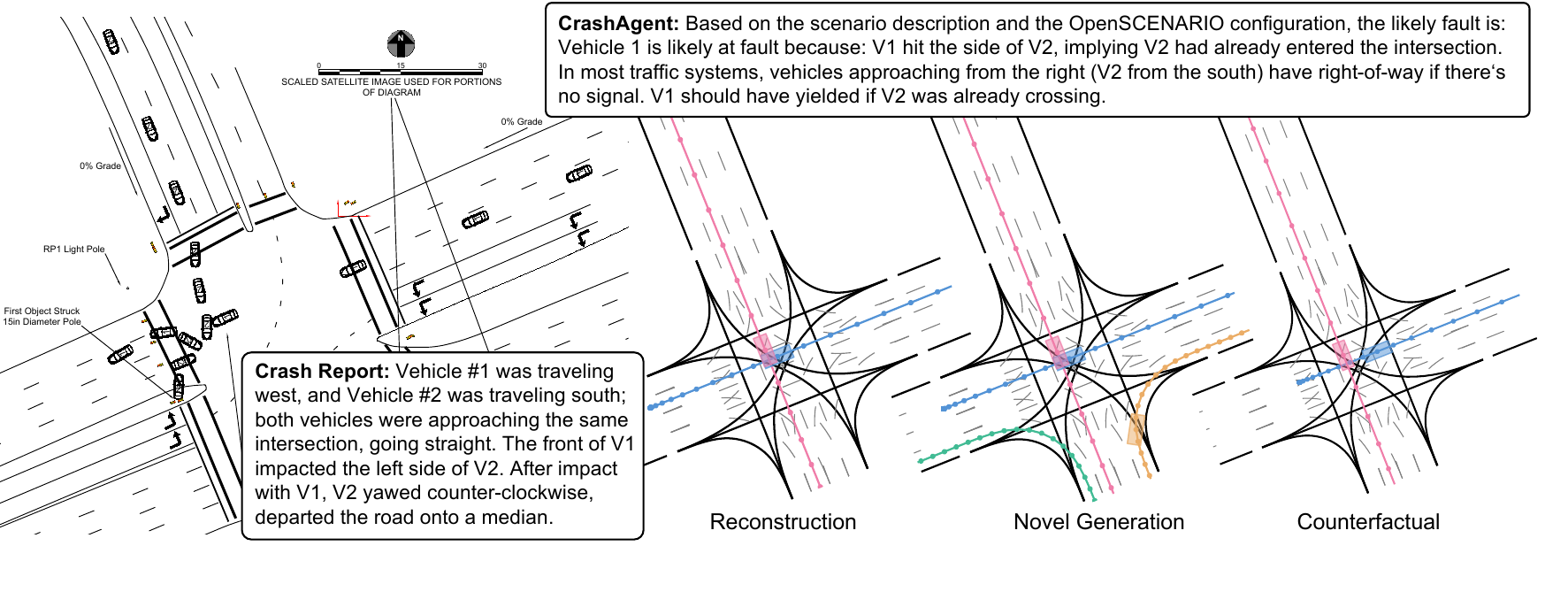}
\vspace{-8mm}
\caption{Generation results from CrashAgent with a crash between two vehicles in an interaction.}
\vspace{-3mm}
\label{fig:example2}
\end{figure}

\textbf{Scenario Reconstruction}\quad
In Figure~\ref{fig:example1}, we show an accident where a vehicle hit a deer that suddenly crossed the road. Given the diagram and textual description, we can see that CrashAgent successfully reconstructs the accident. We also show another example in Figure~\ref{fig:example2} where two vehicles collide in an intersection. CrashAgent reconstructs the multi-lane road network and the collision that is similar to the crash diagram.

\textbf{Novel Scenario Generation}\quad
Based on the crash scenario, we can even generate novel safety-critical scenarios that do not exist in the crash dataset. Specifically, we use the reconstructed scenarios as references and add new NPCs to the scenarios. We show two examples in Figure~\ref{fig:example1} and~\ref{fig:example2}, where the novel scenario contains more vehicles.

\textbf{Responsibility Identification}\quad
Based on the reconstructed scenario, CrashAgent can further analyze the reason for the accident and identify the responsibility. In the examples of Figure~\ref{fig:example1} and~\ref{fig:example2}, we show the responsibility analysis from CrashAgent. With this analysis, we can generate a counterfactual scenario where the collision is avoided by changing the behavior of the ego vehicle.


\vspace{-2mm}
\section{Conclusion}
\label{sec:conclusion}
\vspace{-2mm}

In this paper, we introduced CrashAgent, a novel framework for generating safety-critical driving scenarios by leveraging the reasoning capabilities of multi-modal large language models. Motivated by the long-tail distribution of real-world driving data, our approach addresses the scarcity and limited diversity of critical events by converting textual and visual crash reports into structured, simulation-ready scenarios.
CrashAgent integrates three specialized agents -- Sketch Agent, Road Agent, and Scenario Agent -- to transform high-level crash descriptions into detailed road layouts and behavior scripts. Through extensive experiments, we demonstrate that the generated scenarios exhibit superior realism, behavioral diversity, and layout accuracy. To support broader research in autonomous driving safety, we release a large-scale, high-quality crash scenario dataset based on real-world accidents. We believe this work offers a scalable and interpretable pathway to improve AV safety by incorporating rare but crucial corner cases into the development cycle.

\clearpage
\section*{Limitation}
\label{sec:limitation}

Despite the strengths of CrashAgent in reconstructing diverse and realistic safety-critical scenarios, several limitations still exist. 
First, the framework relies heavily on the quality and coverage of crash reports, which may vary significantly in detail and clarity. Incomplete descriptions or missing diagrams can lead to ambiguous or erroneous scenario reconstructions. 
Moreover, the available reports may not comprehensively represent the full spectrum of rare driving edge cases, introducing a bias toward certain types of accidents.
Additionally, the reconstructed scenarios are designed for use in simulation environments, but differences in simulator assumptions -- such as traffic rules or dynamic models -- can lead to discrepancies between intended and executed behaviors. 
Finally, although VLMs provide strong multi-modal reasoning capabilities, they remain prone to hallucination and may generate logically inconsistent or physically implausible outcomes, particularly in complex scenes. 
These limitations suggest opportunities for future improvements in VLM spatial understanding, model fine-tuning, and simulator integration.

\acknowledgments{}


\clearpage
\appendix

\section{NHTSA Crash Report Dataset}

\subsection{Data Statistics}

\begin{figure}[h]
\centering
\includegraphics[width=1.0\linewidth]{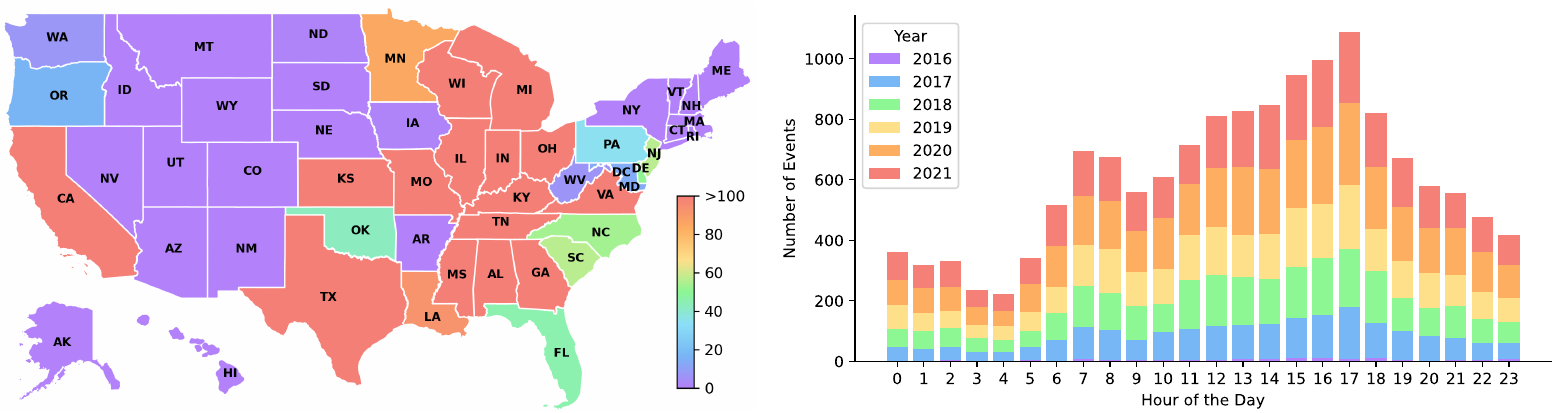}
\caption{The distributions of the location and time of the crash scenarios in the report.}
\label{fig:location_time}
\vspace{-5mm}
\end{figure}

The crash reports are collected from the United States and stored on the CISS website. We plot the distribution of the location of all these crashes and the distribution of the occurrence time in Figure~\ref{fig:location_time}. According to the occurrence time distribution, most crashes happen during the evening rush hours.

\subsection{Data Pre-processing}

The NHTSA crash report also contains metadata about the accidents. We apply rule-based pre-processing to extract information such as the orientation of road users and roads, groups of users originating from the same direction or lane, pairs of users in adjacent lanes, posted speed limits, and the presence of stop signs. This extracted information is summarized in natural language and supplied as supplemental input to the VLM.

\section{Responsibility Analysis and Counterfactual Verification}
\label{app:responsiblity}

In addition to synthesize scenarios that carry the critical factors in the traffic accident reports, we also prompt the LLM with the detailed scenario description and the generated OpenScenario script to return the responsibility analysis.
And in addition to replaying the generated scenario in a simulator to verify that the crash occurs as described, we also implement counterfactual scenarios to validate the responsibility analysis performed by the LLM. 

Specifically, we invoke LLM to either add one pair of trigger and action or identify the faulty trigger-action pair of the initiating road user in the OpenScenario script. We then adjust the corresponding trigger and action parameters along the direction suggested by the LLM, searching until we obtain parameter values that prevent the crash. The absence of crash in the counterfactual scenario provides evidence that this trigger-action pair contributes to the original accident.
We acknowledge that modifying the responder’s behavior can likewise prevent the collision. 
However, as our parameter‑replacement toolbox does not encode traffic regulations or legal right‑of‑way, such modifications alone are insufficient to identify the initiator. 
Instead, we employ counterfactual simulation as an auxiliary validation mechanism for the responsibility assessment. Examples of responsibility analysis are shown below. The initiators are identified correctly.

\begin{tcolorbox}[
  colback=white,       
  colframe=black,      
  boxrule=0.5pt,       
  arc=2mm,             
  left=4pt,            
  right=4pt,           
  top=4pt,             
  bottom=4pt           
]
\textbf{Example 1}
{\small
The collision happened when “the right side of Vehicle 1 contacted the front of Vehicle 2,” \\
indicating that Vehicle 1 had partially entered the intersection before being struck by Vehicle 2. \\
Vehicle 2 appears to be at fault because:\\
The scenario shows both vehicles arriving at the intersection at approximately the same time. \\
Based on the collision point (right side of Vehicle 1 hit by front of Vehicle 2), \\
Vehicle 1 had established presence in the intersection first. \\
Vehicle 2 failed to yield to the vehicle already in the intersection
}
\end{tcolorbox}

\begin{tcolorbox}[
  colback=white,       
  colframe=black,      
  boxrule=0.5pt,       
  arc=2mm,             
  left=4pt,            
  right=4pt,           
  top=4pt,             
  bottom=4pt           
]
\textbf{Example 2}
{
\small
Based on the scenario description and the OpenSCENARIO configuration:\\
Crash Summary:\\
Vehicle 1 (V1) is traveling west.\\
Vehicle 2 (V2) is traveling south.\\
Both vehicles are going straight and approach the same intersection.\\
V1 impacts the left side of V2. This suggests V1 hit V2 in the intersection.\\
V2 then yawed counter-clockwise, left the road, and struck a traffic pole.\\
Likely Fault:\\
Vehicle 1 is likely at fault because:\\
V1 hit the side of V2, implying V2 had already entered the intersection. \\
In most traffic systems, vehicles approaching from the right (V2 from the south) 
have right-of-way if there's no signal.
V1 should have yielded if V2 was already crossing.
}
\end{tcolorbox}

\section{Sketch Agent}
\label{app:sketch_agent}

\subsection{Details of Visual-Tree}
As described in the Visual Tree‑of‑Thought Prompting section, we generate four types of visual trees to decompose a sketch into progressively finer subregions, where the simplest ones only show one region of interest or the least number of components.
To achieve this, the trees are built with explicit awareness of road edges, lane edges, and road users. 
To obtain these semantic labels, we proceed in four steps.
First, connected components are detected from the sketch image. 
Second, Each extracted component is presented to the vision‑language model (VLM), which returns one of six labels: road edge, lane edge, road user, traffic sign, annotation, or other.
Third, we present the components that contain road users to the VLM again with higher resolution, which returns the id of road users based on the label shown in the image.
Finally, with these classification labels, we extract the region of interest and build the trees.

\begin{figure}[!ht]
    \centering
    \begin{subfigure}[b]{0.3\textwidth}
        \centering
        \includegraphics[width=\textwidth]{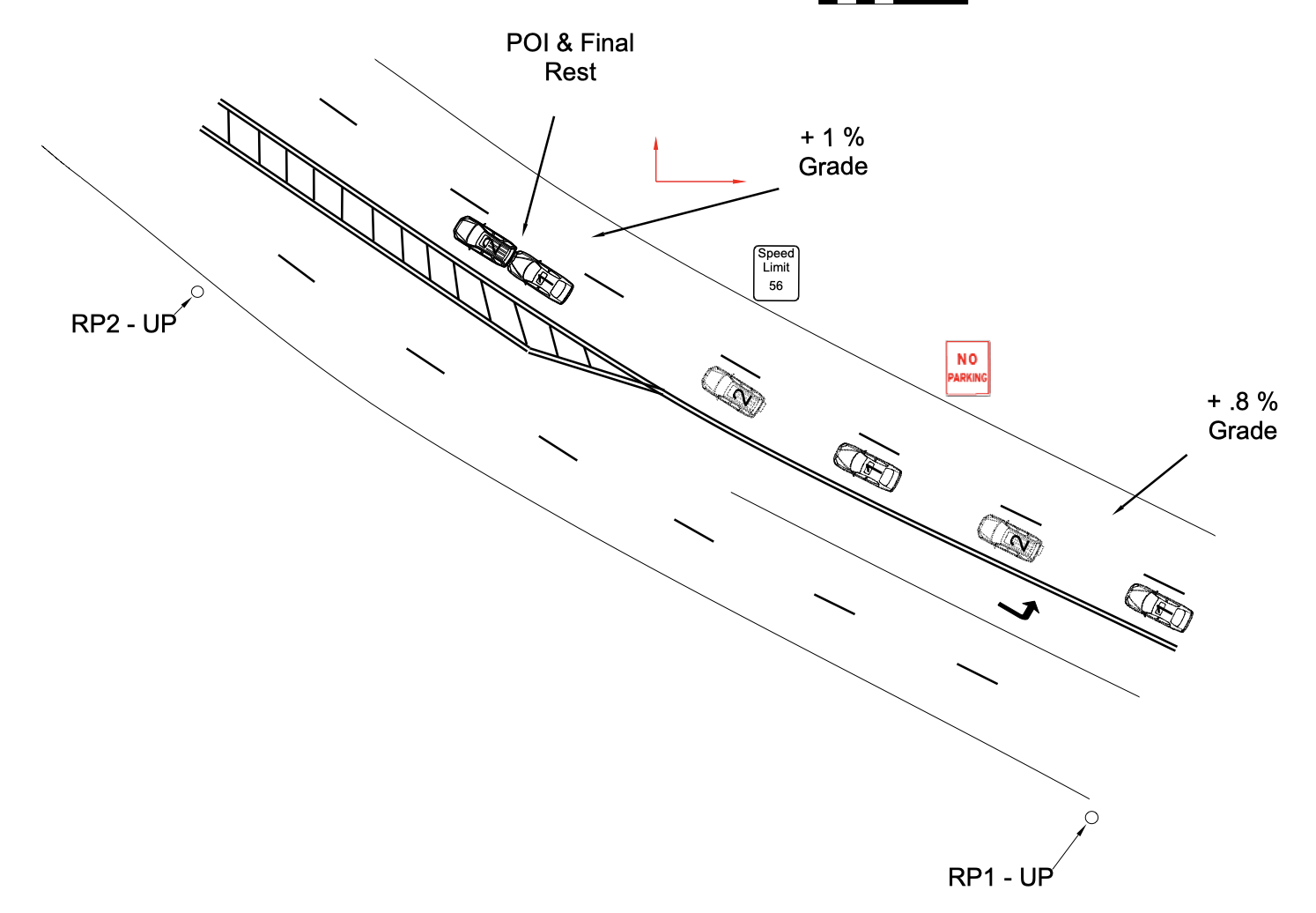}
        \caption{Original Sketch image.}
        \label{fig:sub1}
    \end{subfigure}
    \hfill
    \begin{subfigure}[b]{0.34\textwidth}
        \centering
        \includegraphics[width=\textwidth]{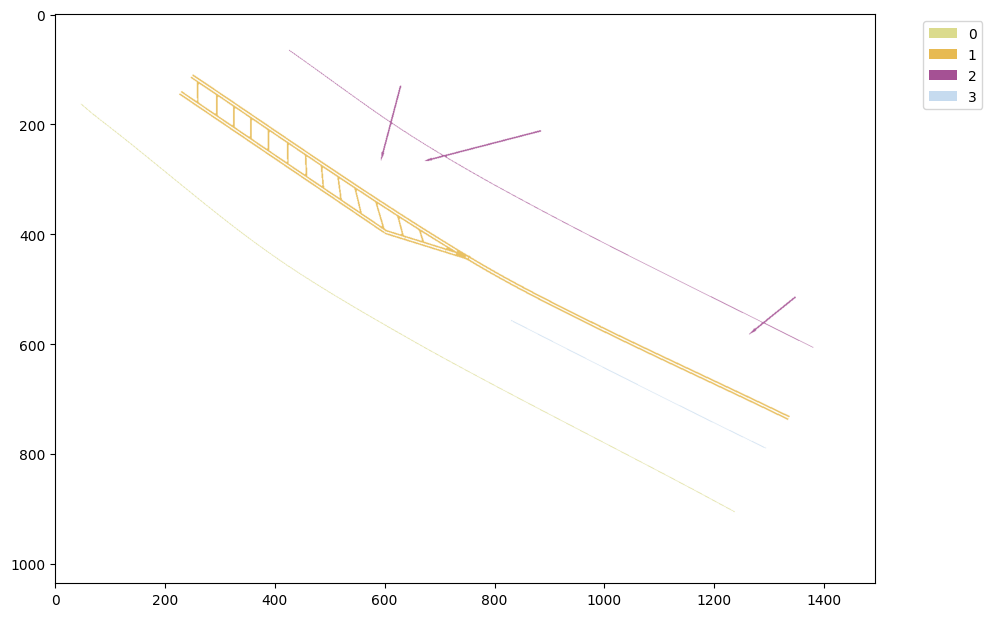}
        \caption{Extracted major road edges.}
        \label{fig:sub2}
    \end{subfigure}
    \hfill
    \begin{subfigure}[b]{0.34\textwidth}
        \centering
        \includegraphics[width=\textwidth]{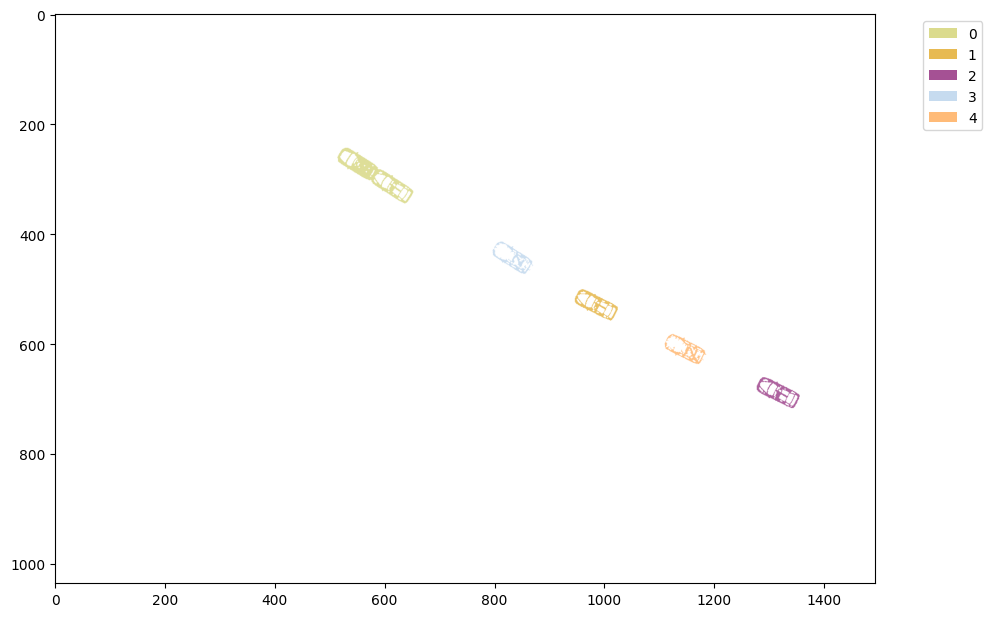}
        \caption{Extracted road users.}
        \label{fig:sub3}
    \end{subfigure}
    \caption{Example of connected component extraction and classification.}
    \label{fig:connected components}
    \vspace{-4mm}
    
    
\end{figure}

An example of  connected component extraction and classification is shown in Figure~\ref{fig:connected components}. In Figure~\ref{fig:sub1}, the original sketch image shows a two-way single road with multiple lanes and multiple vehicle icons showing the scenario. In Figure~\ref{fig:sub2}, the detected major road edges are shown. Based on their position, we use a rule-based method to group them into road boundary pairs and lane edges. In Figure~\ref{fig:sub3}, the detected road users are shown. Note that when multiple vehicle icons contact, they are recognized as one component, and this component is registered for both road users.

\begin{figure}[!ht]
    \centering
    \includegraphics[width=0.8\linewidth]{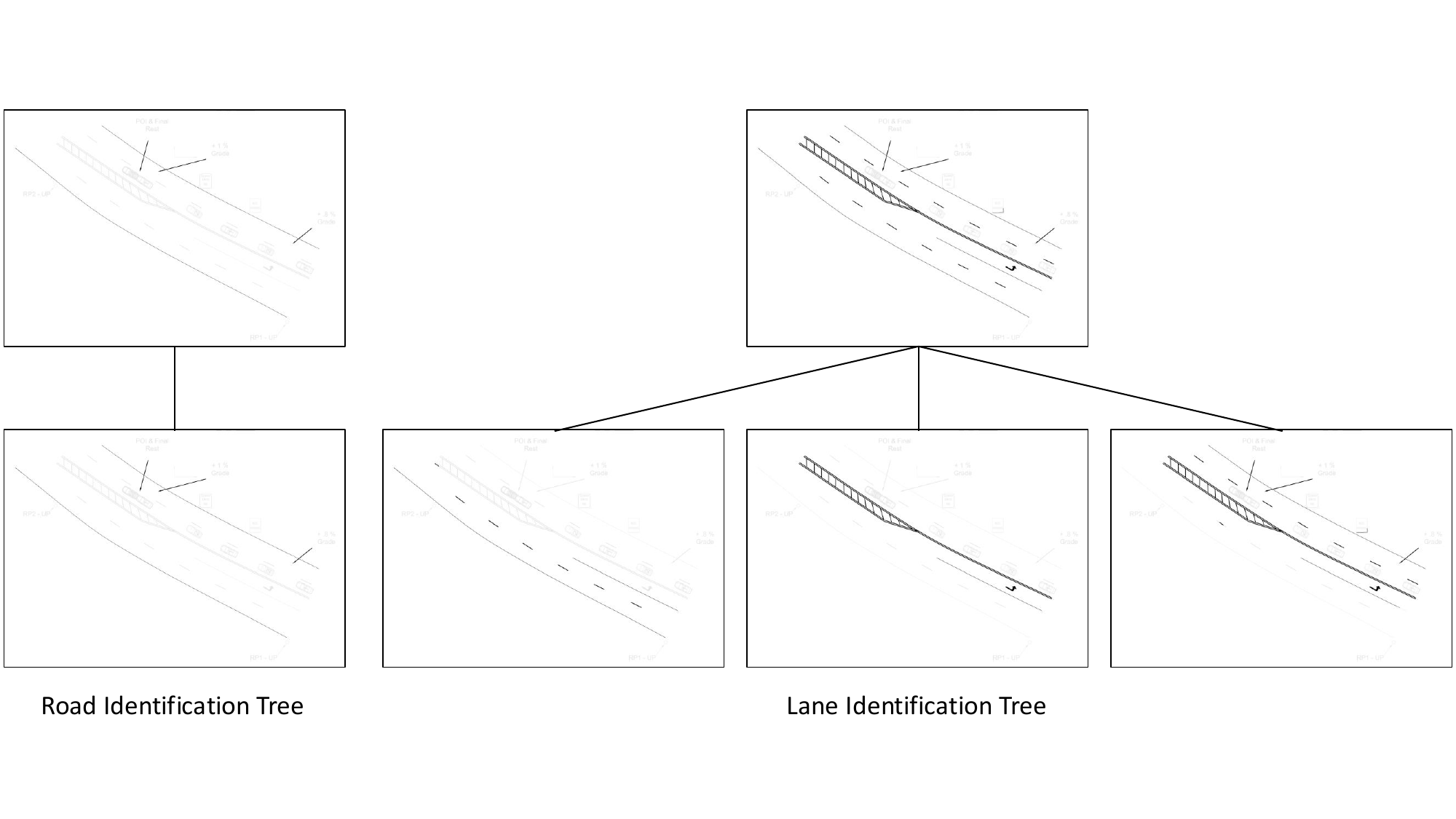}
    \vspace{-7mm}
    \caption{Road Identification Tree and Lane Identification Tree.}
    \label{fig:road tree}
\end{figure}

\begin{figure}[!ht]
    \centering
    \includegraphics[width=0.8\linewidth]{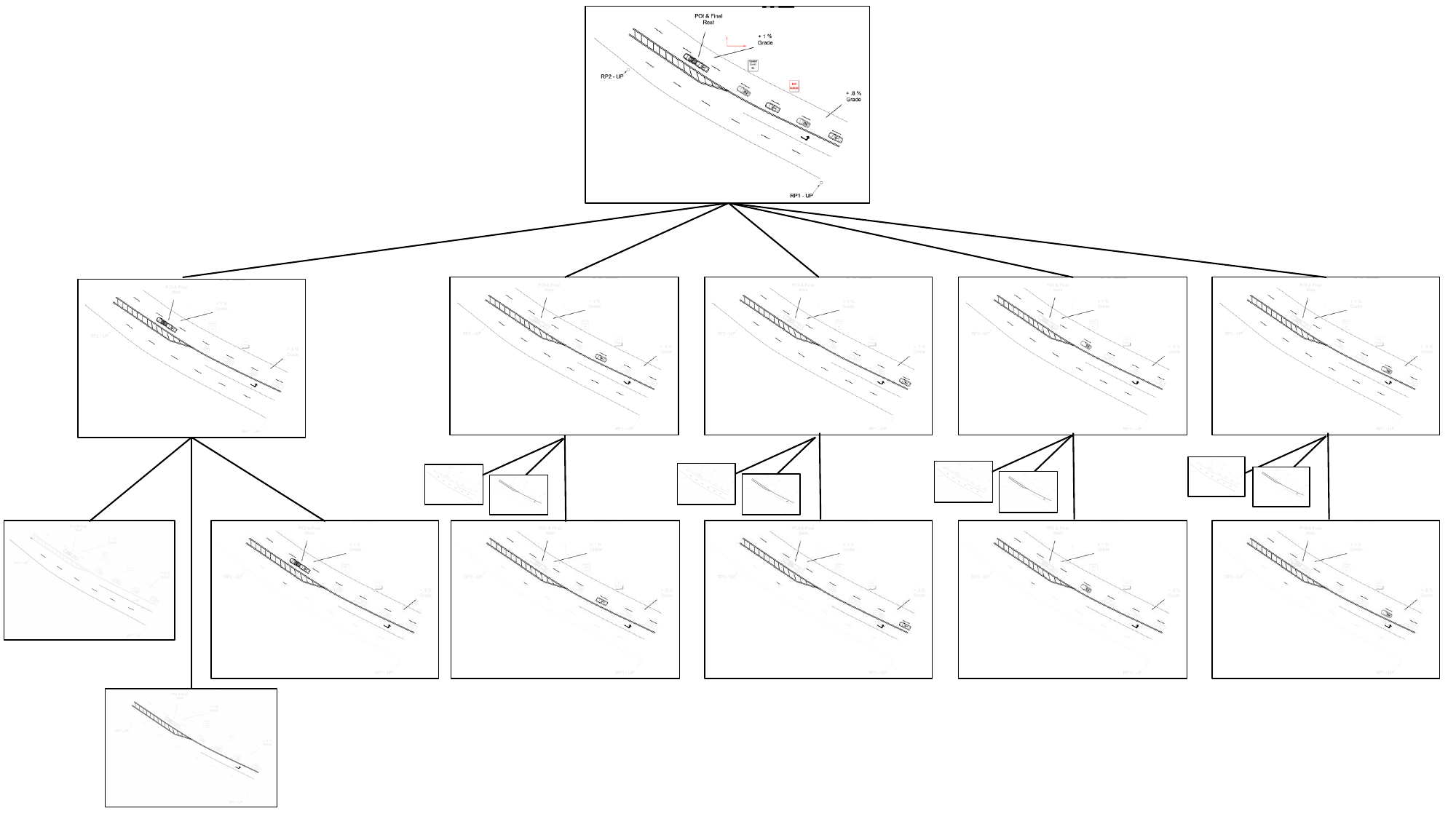}
    \vspace{-7mm}
    \caption{Road User Location Tree}
    \label{fig:road user location tree}
\end{figure}

\begin{figure}
    \begin{subfigure}[b]{1.\textwidth}
        \vspace{-3mm}
        \centering
        \includegraphics[width=0.99\textwidth]{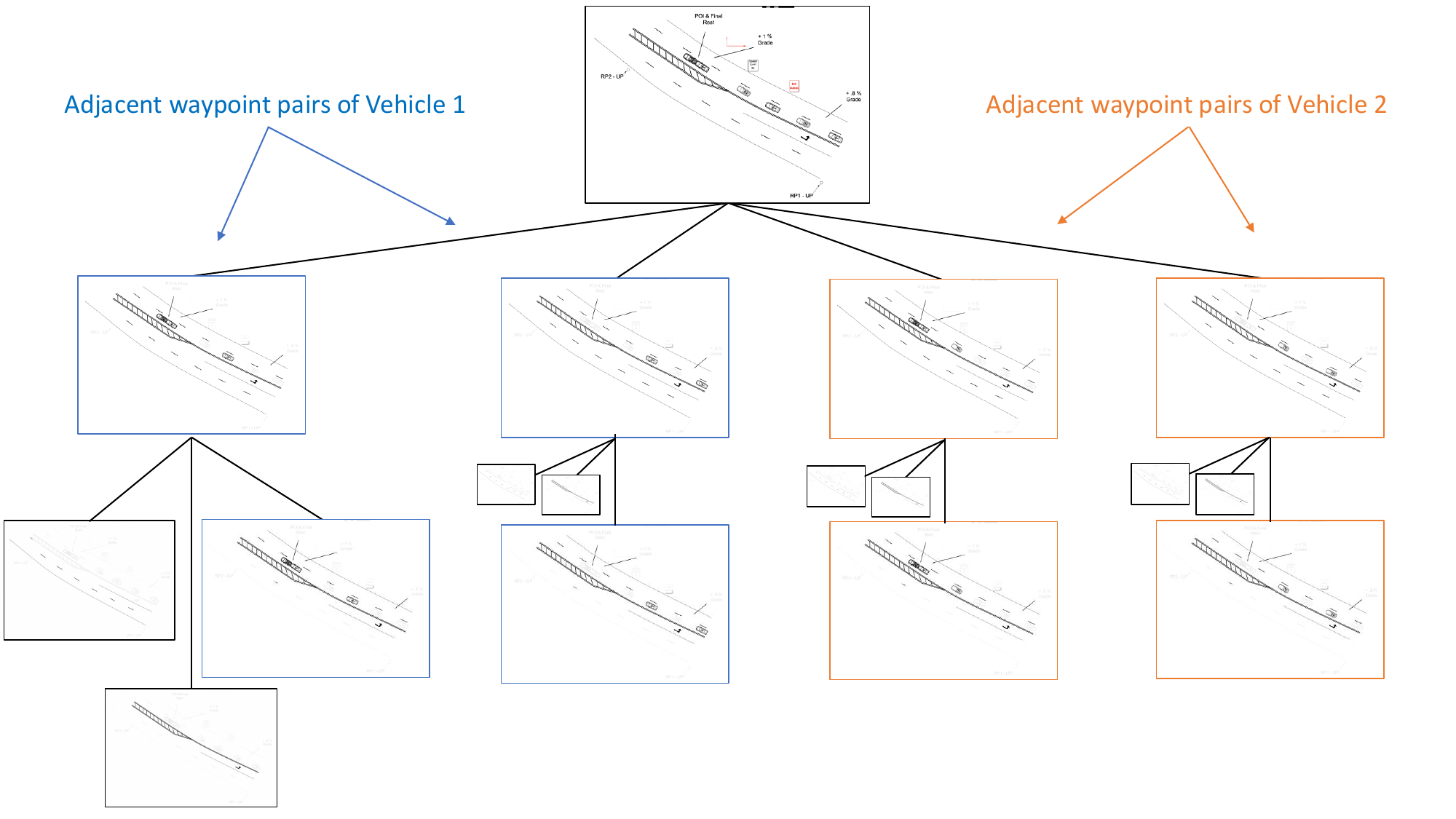}
        \vspace{-2mm}
        \caption{Example of Road User Behavior Tree (Adjacent waypoints).}
        \label{fig:road user behavior tree segment}
    \end{subfigure}
    \vskip\baselineskip
    \begin{subfigure}[b]{1.\textwidth}
        \vspace{-3mm}
        \centering
        \includegraphics[width=0.7\textwidth]{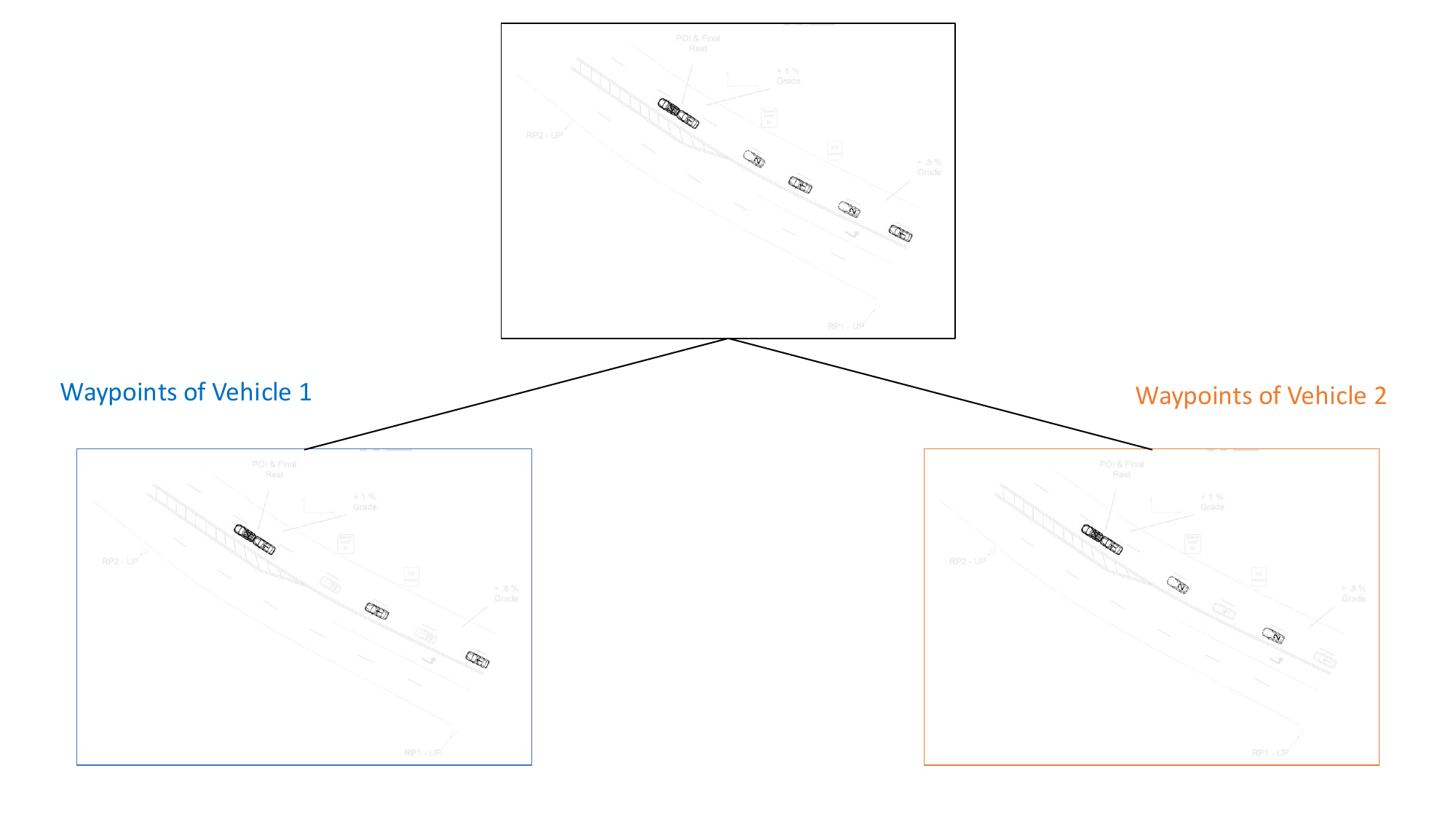}
        \vspace{-3mm}
        \caption{Example of Road User Behavior Tree (Pure waypoints).}
        \label{fig:pure traj tree}
    \end{subfigure}
    \caption{Examples of Road User Location Tree and Road User Behavior Tree.}
    \label{fig:road user behavior tree}
\end{figure}

Figure~\ref{fig:road tree}, Figure~\ref{fig:road user location tree}, and Figure~\ref{fig:road user behavior tree} illustrate an example of our visual trees. 

As shown in Figure~\ref{fig:road tree}, since this is a single‑road scenario, the road identification tree has only one leaf. For the lane identification tree, we partition the image along its primary edges into three subregions, each containing a manageable number of lanes. 

For the road user location tree, as shown in Figure~\ref{fig:road user location tree}, we create one branch for each detected road user icon, anchored to the road on which it appears. For each leaf node, we crop the sketch to the band of that road delimited by its two primary edges. As a result, every leaf shows only a limited number of lanes, enabling the Sketch Agent to identify the location of each user by having the VLM count lanes in that confined region.

For the road user behavior tree, as shown in Figure~\ref{fig:road user behavior tree}, two trees are used. The first tree, as shown in Figure~\ref{fig:road user behavior tree segment}, is created in a similar way to the road user location tree, except that two adjacent waypoints are considered in each branch, while the second tree, as shown in Figure~\ref{fig:pure traj tree}, shows only the waypoints of road users with all road edges removed.

The number of total images used in the Sketch Agent is summarized in Table~\ref{tab:augmentation_levels}. 

\begin{table}[h]
\centering
\vspace{-3mm}
\caption{Number of nodes at each level. $L_i$ = number fo segments in the $i$-th pair of estimated road boundary, $L_j$ = number of segments of in the road where the $j$-th waypoint is located.}
\vspace{1mm}
\label{tab:augmentation_levels}
\resizebox{0.95\linewidth}{!}{
\begin{tabular}{l|c|c|c|c|c}
    \toprule
    \multirow{2}{*}{\textbf{Level}} 
      & \textbf{Road} 
      & \textbf{Lane} 
      & \textbf{Road User Location} 
      & \multicolumn{2}{c}{\textbf{Road User Behavior}} \\
    \cmidrule(lr){2-2} \cmidrule(lr){3-3} \cmidrule(lr){4-4} \cmidrule(lr){5-6}
     & group 1 & group 1 \& 2 & group 1  & group 1  & group 2 \\
     \midrule
     root & 1       & 1     & 1     & 1     & 1 \\
     mid & na       & $R=$\#Road     & $\#$Waypoints    & $W-V=~$\#Waypoints-\#RoadUsers     & na \\
     leaf  & $\#$Road    & $\sum_{i \in [R]} (L_i+1)$ 
                      & $\sum_{j \in [W]} (L_j+1)$ 
                      & $\sum_{j \in [W-V]} (L_j+1)$  & V\\
    \bottomrule
\end{tabular}
}

\end{table}





\section{Road Agent}
Examples of VLM response in Road Agent are shown below.
{
\tiny
\begin{lstlisting}
[Category]:
    Interchange
[Topology]:
    Off Ramp
[Lane info]:
    [2 | 2]
[Road curvature]:
    0.1
[Angle to junction]:
    [N/A]
[Ramp info]:
    [None | Off-ramp]
[Explanation]:
    The diagram shows a single road with a sharp curve, indicating an off-ramp. 
    The crash summary describes V1 traveling southbound on an off-ramp, which matches the road layout. 
    The road has two lanes in each direction, and the curvature is positive, indicating a rightward curve. 
    The ramp type is an off-ramp on the right side, with no on-ramp on the left side.

[Category]:
    Intersection
[Topology]:
    T-Junction
[Lane info]:
    road1: [2 | 3]
    road2: [2 | 3]
    road3: [2 | 2]
[Road curvature]:
    road1: 0
    road2: 0
    road3: 0
[Angle to junction]:
    road1: 0
    road2: pi
    road3: 3pi/2
[Ramp info]:
    [N/A]
[Explanation]:
    The image shows a T-junction with three roads. 
    The main road has five lanes (two left lanes and three right lanes) 
    and the side road has four lanes (two left lanes and two right lanes). 
    The crash summary indicates that the vehicles were traveling south in the right lane of a five-lane roadway, 
    which matches the main road in the T-junction. 
    The roads are straight, so the curvature is 0 for all roads. 
    The angles to the junction are 0 for the north-pointing road, 
    pi for the south-pointing road, and 3pi/2 for the east-pointing road.

[Category]:
    Intersection
[Topology]:
    4-way
[Lane info]:
    road1: [2 | 2]
    road2: [2 | 2]
    road3: [2 | 2]
    road4: [2 | 2]
[Road curvature]:
    road1: 0
    road2: 0
    road3: 0
    road4: 0
[Angle to junction]:
    road1: 0
    road2: pi/2
    road3: pi
    road4: 3pi/2
[Ramp info]:
    [N/A]
[Explanation]:
    The image shows a 4-way intersection with four roads converging at a junction point. 
    Each road has two lanes in each direction, making the lane info [2 | 2] for all roads. 
    The roads are straight, so the road curvature is 0 for all roads. 
    The angles to the junction are determined based on the ...
    The crash summary indicates that Vehicle #1 was traveling East (road4) and Vehicle #2 was traveling North (road1), 
    confirming the intersection type and road layout.
\end{lstlisting}
}

\begin{table}[!ht]
\centering
\caption{Scenario elements}
\vspace{1mm}
\label{tab:scene_template}
\resizebox{1.0\linewidth}{!}{
\begin{tabular}{c|p{5cm}|p{9cm}|p{9cm}}
\toprule
Index & Element Name & Description & Parameters   \\
\midrule
1  & Initial Pose &    set new initial positions for road users so that they're in right order    &    old positions, order of road users, relative distances, road length   \\

2  & Initial Route &   use key waypoints in LanePosition format to assign a route for the given road user which is activated at timestamp 0   &  road user id, key waypoints in LanePosition format   \\

3  & Initial Trajectory &   use key waypoints in LanePosition format and timestamps to assign a trajectory for the given road user which is activated at timestamp 0     &   road user id, key waypoints in LanePosition format, timestamps, timestamp scale ratio    \\

4  & Initial Route World Position &   use key waypoints in WorldPosition format to set up a route for the given road user which is activated at timestamp 0       &    road user id, key waypoints in WorldPosition format    \\

5  & init trajectory world position &   use key waypoints in WorldPosition format and timestamps to set up a trajectory for the given road user which is activated at timestamp 0     &   road user id, key waypoints in WorldPosition format, timestamps, timestamp scale ratio   \\

6  & speed change and init route &    set speed change at timestamp t and use initial and destination position in LanePosition format to assign a route for the given road user which is activated at timestamp t   &   road user id, target speed, time to change speed, distance to change speed, initial and destination position in LanePosition format   \\

7  & speed change and init trajectory &   set speed change at timestamp t and use initial and destination position in LanePosition format to assign a trajectory for the given road user which is activated at timestamp t     &   road user id, target speed, time to change speed, distance to change speed, initial and destination position in LanePosition format, timestamps, timestamp scale ratio     \\

8  & speed change and init route world position &   set speed change at timestamp t and use initial and destination position in WorldPosition format to assign a route for the given road user which is activated at timestamp t   &   road user id, target speed, time to change speed, distance to change speed, initial and destination position in WorldPosition format \\

9  & lane change at time &   set lane change action for road user at timestamp t     &   road user id,  target lane id(s),  timestamp(s) to trigger lane change   \\

10 & lane change at traveled distance &   set lane change within distance $d_dyn$ action for road user at traveled distance $d$   &   road user id(s),  target lane id(s),  traveled distance(s) to trigger lane change and transition distance to complete lane change  \\

11 & lane change to pos &    set a series of lane change actions and make the last lane change action end at target position, while keeping safe distance from other road users   &  road user id, surrounding road user id(s), start position, target position, lane change distance(s)     \\

12 & lane change and speed change at t &   set combination of speed change and lane change actions at corresponding timestamps    &    road user id, target speed, target lane id, timestamps, distance to complete the action, flag indicating if speed change or lane change happens first   \\

13 & parking pulling in &   set pulling-in action for road user while keeping a safe distance from other road users     &    road user id, other road user id(s), current position, parking spot position, distance ranges for pulling-in adjustment, time intervals between pulling-in steps  \\

14 & parking pulling out &   set pulling-out action for road user while keeping a safe distance from other road users     &    road user id, other road user id(s), current position, parking spot position, distance ranges for pulling-out adjustment, time intervals between pulling-out steps  \\

15 & parking pulling over &   set pulling-over action for road user while keeping a safe distance from other road users     &    road user id, other road user id(s), current position, parking spot position, distance ranges for pulling-over adjustment   \\

16 & speed change at t &   set speed change at timestamp t for road user     &   target speed(s), timestamp(s) to trigger speed change action, distance(s) to complete speed change    \\

17 & speed change at traveled distance &   set speed change conditioned on traveled distance d for road user     &   target speed(s), traveled distance(s) to trigger speed change action and to complete speed change  \\

18 & speed change at distance from junction &    set speed change action conditioned on distance to the junction    &   road user id, initial/current position,  initial speed, target speed(s), distance threshold(s)   \\

19 & speed change at distance from junction with traveled distance &   set speed change action conditioned on both traveled distance and distance to the junction    &   road user id, initial/current position,  initial speed, target speed(s), distance to junction threshold(s), traveled distance threshold(s)     \\

20 & offset                         &    set offset action at timestamp t for road user    &   road user id, target offset in lane, timestamp    \\

21 & crash and one stop             &    set stop action for one road user after it collides with another road user    &   road user id, other road user id, time interval to come to stop    \\

22 & crash and stop                 &   set stop action for both road users after collision     &   2 road user ids, 2 time intervals to come to stop    \\

23 & crash and move                 &   set teleport action for both road users after collision     &   2 road user ids, 2 positions in LanePosition format after the collision    \\

24 & crash and lane change          &   set lane change action for road user after it collides with another road user    &   road user id, other road user id, target lane id, time interval to complete lane change   \\

25 & crash and speed change         &   set speed change action for road user after it collides with another road user    &   road user id, other road user id, target speed, time interval to complete speed change  \\

26 & stop at stop sign              &   set speed change action for road user such that it stops at the stop sign    &   road user id, current position, distance to start decelerate     \\
27 & start at stop sign             &   set speed change action for road user such that it accelerates at the stop sign and keeps a safe distance from other road users after waiting for a while   &   road user id, current position, distance to start decelerate     \\

28 & accelerate to cross intersection &     set acceleration speed change action conditioned on distance to the junction    &   road user id, intersection position,  target speed, distance threshold         \\

29 & decelerate before cross intersection &    set deceleration speed change action conditioned on distance to the junction    &   road user id, intersection position,  target speed, distance threshold         \\

30 & stop and wait until another entity cross intersection &    set speed change actions for road user so that it waits for the other road user to cross the intersection and then accelerates to cross the intersection   &  road user id, the other road user id, target speed to cross, 2 distance thresholds \\

31 & stop and cross intersection with another entity &    set speed change actions for road user so that it waits for the other road user to approach the intersection and then accelerates to cross the intersection within a distance threshold to the other road user  &  road user id, the other road user id, target speed to cross, 2 distance thresholds   \\

32 & close lane change and immediate crash &   set lane change action for road user when it's close to another road user so that they collide     &    road user id, other road user id, target relative speed, relative distance   \\

33 & close lane change and immediate avoidance &    set lane change actions and speed change actions to complete the scene where 1) one road user to adopt lane change when it's close to another road user, 2) the road user does lane change again to avoid collision, 3) the road user ends up off-road and stops   &    road user id, other road user id, target relative speed, relative distance threshold   \\

34 & start from ramp and merge      &    set speed change to follow the traffic on the main road    &   road user id, other road user id, distance threshold, relative speed   \\

35 & offset at position             &   set offset action when approaching specific position     &    road user id, offset value, target position, distance threshold  \\

36 & offset with other entity       &   set offset action when approaching another road user     &    road user id, other road user id, distance threshold, offset value    \\

37 & aware of risk and speed change &   set speed change when it is too close to another road user     &    road user id, other road user id, distance threshold, target speed   \\

38 & aware of risk and lane change &    set lane change when it is too close to another road user     &    road user id, other road user id, distance threshold, target lane id   \\

39 & aware of risk and offset      &    set offset action when it is too close to another road user     &    road user id, other road user id, distance threshold, target offset value    \\

40 & aware of risk and lose control and stop &   set lane change and speed change actions to implement when getting too close to another road user, running off-road and coming to stop scene     &  road user id, other road user id, distance threshold, direction to run off-road, time to come to stop     \\

41 & aware of risk and turn traj   &   set new trajectory waypoints and timestamps for road user when it is too close to another road user and tries to avoid collision    &   road user id, other road user id, distance threshold, original trajectory waypoints and timestamps, avoidance direction and distance   \\

42 & avoid and secondary crash     &   set distance action such that when the road user tries to avoid collision with one road user, it collides with another road user     &   road user id, id of the road user to avoid, id of the road user to collide with, distance threshold, time interval to crash    \\
\bottomrule
\end{tabular}
}
\end{table}

\section{Scenario Agent}

\subsection{Details of Scenario Elements}
\label{app:scene_template}

We define 42 scenario elements as building blocks to compose crash scenarios. These elements are summarized in Table~\ref{tab:scene_template}. 

\subsection{Semantic Level Description}
Instead of invoking VLM to emit an OpenScenario script directly, the proposed Scenario Agent first prompts the VLM to return a dictionary of high-level semantic descriptions for both road users and events. These descriptions are then translated into a sequence of primitive scenes selected from the 42 scenario elements. Examples of the VLM response are shown below.
Despite the various keys, every response adheres to a two‑part structure, namely the road user description and the event description.
We use the road user description to assign routes, trajectories, and other behaviors that establish the intent of each road user. 
We use the event description to select the primitive scenes necessary to reconstruct single entity event, their interactions, and the post-crash status.

\textbf{Example 1: Vehicles collide when crossing intersection}
{
\tiny
\begin{lstlisting}
{
    'number of vehicle': 2,
    'number of crash event': 1,
    'number of road segment branch': 4,
    'driving vehicle desc': [
        {
            'name': 'Vehicle 1',
            'state': 'normal driving',
            'init lane': 'right most lane',
            'accident lane': 'right most lane',
            'action/reaction': ['proceed', 'crash'],
            'action/reaction reason': ['normal drive: proceeding', 'crash'],
            'final state': 'stop on road',
            'involved in single-vehicle crash': False,
            'involved in crash with vehicle': True,
            'involved in crash with stopped/parked/pulling-in/pulling-out vehicle': False,
        },
        {
            'name': 'Vehicle 2',
            'state': 'normal driving',
            'init lane': 'right most lane',
            'accident lane': 'right most lane',
            'action/reaction': ['proceed', 'crash'],
            'action/reaction reason': ['normal drive: proceeding', 'crash'],
            'final state': 'stop on road',
            'involved in single-vehicle crash': False,
            'involved in crash with vehicle': True,
            'involved in crash with stopped/parked/pulling-in/pulling-out vehicle': False,
        }
    ],
    'queue on the road': []
}

{
    'number of vehicles': 2,
    'traffic control': {'Vehicle 1': 'none', 'Vehicle 2': 'none'},
    'init location': {'Vehicle 1': 'west', 'Vehicle 2': 'south'},
    'movement': {'Vehicle 1': 'go straight', 'Vehicle 2': 'go straight'},
    'lane change summary': {},
    'init velocity': {'Vehicle 1': 11.11, 'Vehicle 2': 13.33},
    'later velocity': {'Vehicle 1': 11.11, 'Vehicle 2': 13.33},
    'involvement': [
        {
            'involved entities': ['Vehicle 1', 'Vehicle 2'],
            'involved entity relative position/posture': ['orthogonal', 'orthogonal'],
            'involved entities state': ['moving forward', 'moving forward'],
            'involvement type': 'crash',
            'crash type': 'basic crash',
            'noncontact involvement type': None,
            'crash style': 'side hit (orthogonal, orthogonal)',
            'single entity action': [],
            'reaction': [['slow down', 'stop'], ['slow down', 'stop']],
            'action/reaction reason': ['avoid vehicle', 'avoid vehicle']
        }
    ]
}
\end{lstlisting}
}

\textbf{Example 2: Single road case where a vehicle changed lane and hit a parked vehicle}
{\tiny
\begin{lstlisting}
{
    'number of moving vehicle': 1,
    'number of parked/parking vehicle': 1,
    'parked/parking vehicle desc': [
        {
            'name': 'Vehicle 2',
            'init lane': 'off-street or outside of road',
            'target lane': 'off-street or outside of road',
            'accident lane': 'off-street or outside of road',
            'state': 'parked or pulled over or stopped',
            'action/reaction': ['crash'],
            'action/reaction reason': ['crash'],
            'final state': 'hit'
        }
    ],
    'driving vehicle desc': [
        {
            'name': 'Vehicle 1',
            'state': 'normal driving',
            'init lane': 'right most lane',
            'accident lane': 'off-street or outside of road',
            'action/reaction': ['proceed', 'turn right', 'crash', 'stop'],
            'action/reaction reason': ['normal drive: proceeding', 'normal drive: lane change', 'crash', 'post-crash'],
            'final state': 'stop offroad',
            'involved in single-vehicle crash': False,
            'involved in crash with vehicle': True,
            'involved in crash with parked/pulling-in/pulling-out vehicle': True,
        }
    ],
    'queue on the road': []
}

[
    {
        'involved entities': ['Vehicle 1', 'Vehicle 2'],
        'involved entities state': ['moving', 'parked'],
        'involvement type': 'crash',
        'crash type': 'basic crash',
        'crash type involving parking/parked vehicle': 'stopped in traffic lane crash',
        'noncontact involvement type': None,
        'crash style': 'parking collision: car following (back, front)',
        'reaction': [['proceed', 'turn right', 'crash', 'stop'], ['crash']],
        'single entity action': None,
        'action/reaction reason': ['normal drive: proceeding', 'normal drive: lane change', 'crash', 'post-crash']
    }
]
\end{lstlisting}
}

\textbf{Example 3: Single road case where a vehicle hit an animal}
{\tiny
\begin{lstlisting}
{
    'number of vehicle': 1,
    'number of animal': 1,
    'number of pedestrian': 0,
    'number of road segment branch': 1,
    'animal or pedestrian desc': [
        {
            'name': 'Animal 1',
            'enter position': 'road',
            'enter side': 'left',
            'cross direction': 'straight',
            'involved in crash': True,
            'reaction': 'proceed',
            'final state': 'hit'
        }
    ],
    'vehicle desc': [
        {
            'name': 'Vehicle 1',
            'reaction': ['proceed'],
            'reaction reason': ['normal drive'],
            'final state': 'stop on road',
            'involved in single-vehicle crash': False,
            'involved in crash with vehicle': False,
            'involved in crash with animal/pedestrian': True,
        }
    ]
}

[
    {
        'involved entities': ['Vehicle 1', 'Animal 1'],
        'involvement type': 'crash',
        'crash type': 'basic crash',
        'noncontact involvement type': None,
        'action': [['proceed', 'crash'], ['proceed', 'crash']],
        'crash style': 'encounter'
    }
]
\end{lstlisting}
}

\textbf{Example 4: Single road user crash}
{
\tiny 
\begin{lstlisting}
{
    'number of vehicle': 1,
    'number of crash event': 1,
    'number of road segment branch': 2,
    'traffic control': {'Vehicle 1': 'none'},
    'intention': {
        'Vehicle 1': 'exit ramp/service road and come to off-road or half off-road position'
    },
    'lane change summary': {
        'Vehicle 1': {
            'lane change direction': ['turn right'],
            'lane change position': ['inside ramp or service road'],
            'lane change reason': ['critical behavior: avoid crash']
        }
    },
    'init velocity': {'Vehicle 1': 31.39},
    'later velocity': {'Vehicle 1': 0},
    'driving vehicle desc': [
        {
            'name': 'Vehicle 1',
            'state': 'normal driving on ramp or service road',
            'init lane': 'left most lane on ramp or service road',
            'accident lane': 'outside of traffic lanes on the left of ramp or service road',
            'accident position': 'on ramp or service road',
            'action/reaction': ['proceed', 'turn right', 'crash'],
            'action/reaction reason': ['normal drive: proceeding', 'critical behavior: avoid crash', 'crash'],
            'final state': 'stop off-ramp or service road',
            'involved in single-vehicle crash': True,
            'involved in crash with vehicle': False,
            'involved in crash with stopped/parked/pulling-in/pulling-out vehicle': False
        }
    ],
    'queue on the road': []
}

[
    {
        'involved entities': ['Vehicle 1'],
        'event position': 'on ramp or service road',
        'event lane': 'outside of traffic lanes on the left of ramp or service road',
        'involved entity relative position/posture': [None],
        'involved entities state': ['normal driving on ramp or service road'],
        'involvement type': 'single entity event',
        'crash type': 'avoid and losing control crash',
        'noncontact involvement type': None,
        'crash style': 'lose control',
        'single entity action': ['proceed', 'turn right', 'crash'],
        'reaction': [],
        'action/reaction reason': ['normal drive: proceeding', 'critical behavior: avoid crash', 'crash']
    }
]
\end{lstlisting}
}

\textbf{Example 5: Highway with ramp or main road with service road case}
{
\tiny
\begin{lstlisting}
{
    'number of vehicle': 2,
    'number of crash event': 3,
    'number of road segment branch': 2,
    'traffic control': {'Vehicle 1': 'none', 'Vehicle 2': 'none'},
    'intention': {
        'Vehicle 1': 'enter highway/main road from ramp/service road',
        'Vehicle 2': 'drive along highway/main road'
    },
    'lane change summary': {
        'Vehicle 1': {
            'lane change direction': ['turn left'],
            'lane change position': ['inside ramp or service road'],
            'lane change reason': ['normal drive: lane change during on ramp or service road']
        },
        'Vehicle 2': {
            'lane change direction': ['turn right'],
            'lane change position': ['inside highway or main road'],
            'lane change reason': ['normal drive: lane change during on highway or main road']
        }
    },
    'init velocity': {'Vehicle 1': 17.78, 'Vehicle 2': 17.78},
    'later velocity': {'Vehicle 1': 0, 'Vehicle 2': 0},
    'driving vehicle desc': [
        {
            'name': 'Vehicle 1',
            'state': 'entering highway or main road from ramp or service road',
            'init lane': 'right most lane on ramp or service road',
            'accident lane': ['right most lane on highway or main road', 'outside of traffic lanes on the right of highway or main road'],
            'accident position': ['on highway or main road', 'on highway or main road'],
            'action/reaction': ['proceed', 'turn left', 'crash'],
            'action/reaction reason': ['normal drive: proceeding', 'normal drive: lane change during on ramp or service road', 'crash'],
            'final state': 'stop on highway or main road with offset in lane',
            'involved in single-vehicle crash': True,
            'involved in crash with vehicle': True,
            'involved in crash with stopped/parked/pulling-in/pulling-out vehicle': False
        },
        {
            'name': 'Vehicle 2',
            'state': 'normal driving on highway or main road',
            'init lane': 'middle lane on highway or main road',
            'accident lane': ['middle lane on highway or main road'],
            'accident position': ['on highway or main road'],
            'action/reaction': ['proceed', 'crash'],
            'action/reaction reason': ['normal drive: proceeding', 'crash'],
            'final state': 'stop on highway or main road in original lane',
            'involved in single-vehicle crash': False,
            'involved in crash with vehicle': True,
            'involved in crash with stopped/parked/pulling-in/pulling-out vehicle': False
        }
    ],
    'queue on the road': [['Vehicle 2', 'Vehicle 1']]
}

[
    {
        'involved entities': ['Vehicle 1', 'Vehicle 2'],
        'event position': 'on highway or main road',
        'event lane': ['right most lane on highway or main road', 'middle lane on highway or main road'],
        'involved entity relative position/posture': ['back', 'front'],
        'involved entities state': ['entering highway or main road from ramp or service road', 'normal driving on highway or main road'],
        'involvement type': 'crash',
        'crash type': 'basic crash',
        'noncontact involvement type': None,
        'crash style': 'car following (back, front)',
        'single entity action': [],
        'reaction': [['crash', 'stop'], ['crash', 'stop']],
        'action/reaction reason': [['crash', 'post crash'], ['crash', 'post crash']]
    },
    {
        'involved entities': ['Vehicle 1'],
        'event position': 'on highway or main road',
        'event lane': ['outside of traffic lanes on the right of highway or main road'],
        'involved entity relative position/posture': [None],
        'involved entities state': ['entering highway or main road from ramp or service road'],
        'involvement type': 'single entity event',
        'crash type': 'lose control',
        'noncontact involvement type': None,
        'crash style': 'lose control',
        'single entity action': ['turn left', 'crash'],
        'reaction': [],
        'action/reaction reason': ['normal drive: lane change during on ramp or service road', 'crash']
    }
]
\end{lstlisting}
}

\end{document}